\documentclass[letterpaper, 10 pt, conference]{ieeeconf}  

\IEEEoverridecommandlockouts                              

\usepackage{graphics} 
\usepackage{epsfig} 
\usepackage{mathptmx} 
\usepackage{times} 
\usepackage{amsmath} 
\usepackage{amssymb}  
\usepackage{url} 
\usepackage{xcolor}
\usepackage{algorithm}
\usepackage{algpseudocode}
\usepackage{pgfplots}
\usepackage{balance}  
\makeatletter
\let\NAT@parse\undefined
\makeatother

\usepackage[hidelinks]{hyperref}
\usepackage{cleveref}  

\title{\LARGE \bf
Learning Dexterous In-Hand Manipulation with Multifingered Hands via Visuomotor Diffusion
}

\author{Piotr Koczy$^{1}$, Michael C. Welle$^{1,2}$, Danica Kragic$^{1}$ 
\thanks{1 Division of Robotics, Perception and Learning (RPL), KTH Royal Institute of Technology, Sweden. {\tt (pkoczy, mwelle, dani@kth.se)}.
}%
\thanks{2 INCAR Robotics AB. {\tt (michael.welle@incar-robotics.se)}.
}%
}

\begin{document}

\maketitle
\thispagestyle{empty}
\pagestyle{empty}

\begin{abstract}

We present a framework for learning dexterous in-hand manipulation with multifingered hands using visuomotor diffusion policies. Our system enables complex in-hand manipulation tasks, such as unscrewing a bottle lid with one hand, by leveraging a fast and responsive teleoperation setup for the four-fingered Allegro Hand. We collect high-quality expert demonstrations using an augmented reality (AR) interface that tracks hand movements and applies inverse kinematics and motion retargeting for precise control. The AR headset provides real-time visualization, while gesture controls streamline teleoperation.
To enhance policy learning, we introduce a novel demonstration outlier removal approach based on HDBSCAN clustering and the Global-Local Outlier Score from Hierarchies (GLOSH) algorithm, effectively filtering out low-quality demonstrations that could degrade performance. We evaluate our approach extensively in real-world settings and provide all experimental videos on the project website. \footnote{\url{https://dex-manip.github.io/}}.

\end{abstract}

\section{Introduction}
\label{sec:introduction}

\begin{figure}
    \centering
    \includegraphics[width=1.0\linewidth]{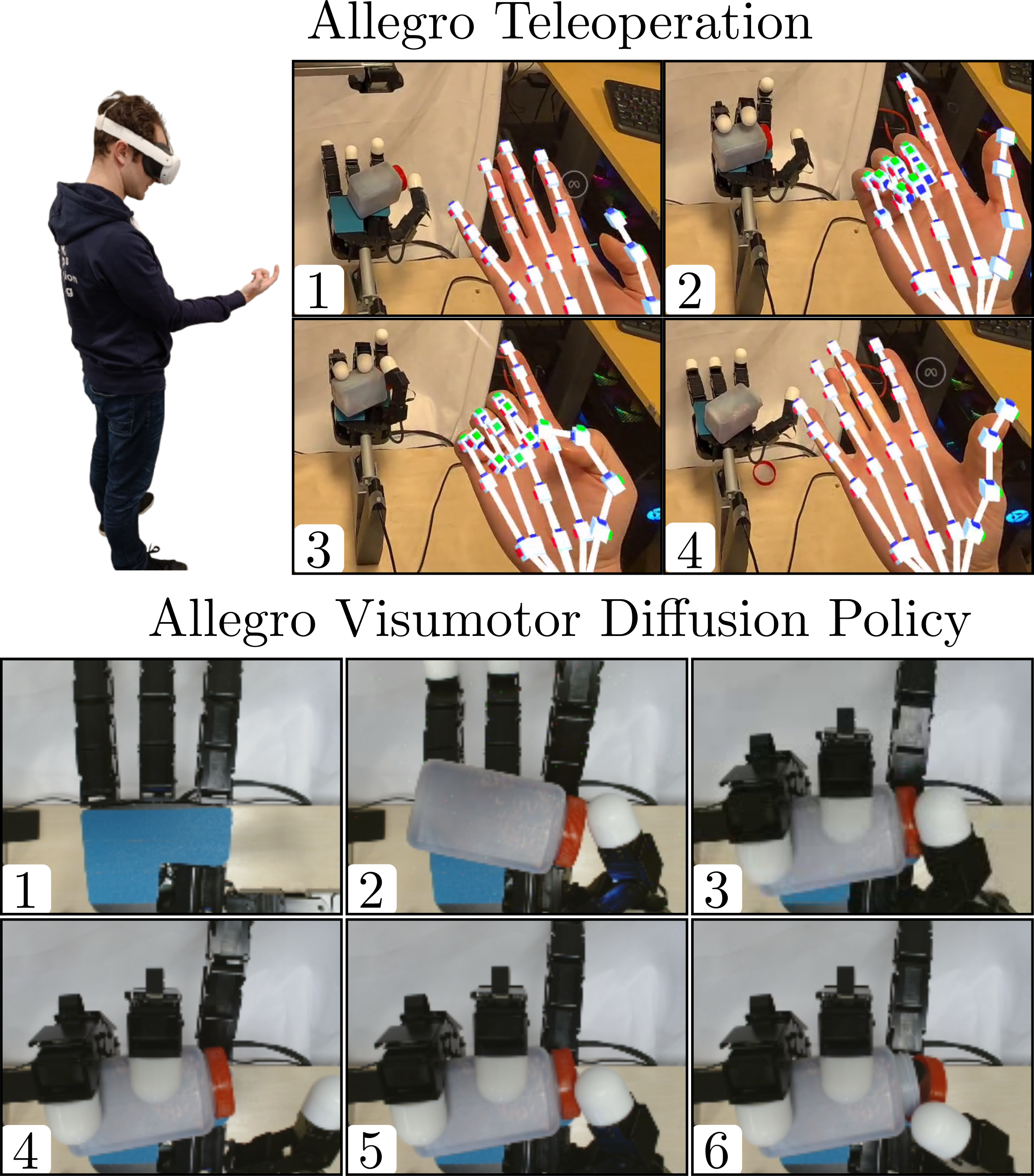}
    \caption{Our Allegro AR teleoperation system on the top shows the operator wearing the AR headset and seeing both the hand tracking and the Allegro hand in view, enabling intuitive and responsive operation.
    On the bottom, we see the trained visuomotor diffusion policy autonomously unscrewing the bottle.
    }
    \label{fig:overview}
\end{figure}

Visuomotor diffusion policies trained on a small number of expert demonstrations have demonstrated mastery in various complex manipulation tasks, such as $6$-DoF mug flipping, sauce pouring, and spreading~\cite{chi2023diffusionpolicy}, opening bottles with a bottle opener and serving rice~\cite{ingelhag2024robotic}, as well as cooking shrimp and wiping wine spills~\cite{fu2024mobile}. In this work, we extend visuomotor diffusion policies~\cite{chi2023diffusionpolicy} to enable complex in-hand manipulation, specifically unscrewing a lid from a container or bottle using a single Allegro Hand.
In-hand manipulation has long been a fundamental challenge in robotics~\cite{salisbury1982articulated, okamura2000overview, andrychowicz2020learning}, as bridging the gap between human and robotic dexterity is crucial for enabling robots to perform everyday tasks with ease. Mastering such fine-grained manipulation is essential for applications in household robotics, industrial automation, and assistive technologies.
To obtain high-quality demonstrations, we developed a hand-tracking and control pipeline, as shown in Fig.~\ref{fig:overview} (top). Our system utilizes an augmented reality (AR) headset for real-time hand tracking, transmitting the detected skeletal graph to a ROS-based processing node. There, we apply a combination of motion retargeting and inverse kinematics to adapt human hand movements to the kinematically different Allegro Hand. This setup allows the operator to teleoperate the Allegro Hand intuitively by simply moving their own hand while wearing the AR headset. The headset operates in passthrough mode, overlaying the operator's real-world view with tracked hand movement visualizations. This real-time feedback loop enhances precision and responsiveness, enabling fine control of in-hand manipulation tasks.

Using this system, we collect $300$ expert demonstrations of the unscrewing task, capturing a diverse range of observables. However, not all demonstrations contribute positively to policy training, as low-quality or outlier attempts can degrade performance. To address this, we integrate an outlier removal algorithm based on HDBSCAN clustering and the Global-Local Outlier Score from Hierarchies (GLOSH), filtering out outlier demonstrations before policy training.
Furthermore, we conduct extensive ablation studies to analyze the impact of different observation modalities. Our results show that combining wrist-mounted camera observations with joint positions and effort readings yields the best performance. This finding highlights the potential of deploying such visuomotor diffusion-based systems on mobile manipulation platforms, such as humanoid robots.

\newpage
Our key contributions are:
\begin{itemize}
    \item An intuitive AR-based teleoperation system for real-time demonstration collection with the Allegro Hand.
    \item An outlier removal strategy using HDBSCAN clustering and GLOSH to improve demonstration quality.
    \item Extensive ablation studies demonstrating that a combination of wrist-camera observations, joint positions, and effort readings leads to the best policy performance for our task.
    \item Experimental validation of visuomotor diffusion policies for in-hand manipulation, underscoring their potential for deployment on mobile manipulation platforms such as humanoid robots.
\end{itemize}

\section{Related Work}
\label{sec:related_work}

\noindent
\textbf{In-Hand Manipulation:}
In-hand manipulation is a fundamental challenge in robotics, requiring precise coordination of fingers and contact forces to dexterously manipulate objects. Early approaches relied on analytic methods and precomputed grasping strategies~\cite{iberall1997human}. More recent advances leverage deep learning and reinforcement learning to enable dexterous manipulation~\cite{andrychowicz2020learning}. While prior works have demonstrated success in reorienting objects~\cite{akkaya2019solving, qin2022dexmv}, few have tackled more complex manipulation tasks such as unscrewing a lid from a bottle in a fully autonomous manner. Our work extends visuomotor diffusion policies to this setting, focusing on leveraging wrist-camera observations alongside proprioceptive signals to enhance policy performance.

\noindent
\textbf{Teleoperation for Dexterous Manipulation:}
Collecting high-quality demonstrations is crucial for training visuomotor policies, and teleoperation provides an effective means for expert data collection. Previous teleoperation systems have utilized exoskeletons~\cite{wei2023adaptive}, motion capture systems~\cite{wang2403dexcap}, and specific tracking sensors~\cite{zahlner2020teleoperation}. However, these approaches often suffer from calibration drift, occlusion issues, or lack of direct feedback to the operator. Our system leverages an AR-based teleoperation setup that provides real-time hand tracking with visual feedback, ensuring intuitive and responsive data collection via inverse kinematics and motion retargeting. This approach enables the collection of high-quality demonstrations for complex in-hand manipulation tasks.

\noindent
\textbf{Diffusion Models for Robotics:}
Diffusion models have recently emerged as a powerful framework for generating complex behavior in robotics, specifically for visuomotor control~\cite{chi2023diffusionpolicy}. This has spawned an array of work such as~\cite{carvalho2023motion, urain2023se, yang2025s, jang2022bc, mees2022matters, xian2023chaineddiffuser, ha2023scaling}, exploring their potential for learning robust and generalizable robotic skills. Unlike reinforcement learning, which requires extensive environment interaction, diffusion models can be trained on offline datasets of expert demonstrations, making them well-suited for robotic manipulation tasks. Our work builds on this paradigm by investigating the role of multimodal sensory input, showing that combining wrist-mounted visual observations with proprioceptive signals leads to superior policy performance.

\noindent
\textbf{Outlier Detection:}
Outlier demonstrations can degrade policy performance, making robust data filtering essential. A comprehensive survey on outlier detection~\cite{boukerche2020outlier} categorizes various approaches, highlighting statistical heuristics, density-based methods, and machine learning-based models for anomaly detection. Building on these insights, we adopt an unsupervised approach using HDBSCAN~\cite{mcinnes2017hdbscan} clustering and the GLOSH~\cite{campello2015hierarchical} algorithm, which effectively detects and removes suboptimal trajectories without requiring explicit supervision. This filtering improves training efficiency and contributes to the overall robustness of the learned policies.

By integrating techniques from these domains, our work advances the state-of-the-art in in-hand manipulation by combining AR-based teleoperation, multimodal visuomotor diffusion policy learning, and demonstration filtering.

\begin{figure*}[h]
    \centering
    \includegraphics[width=\linewidth]{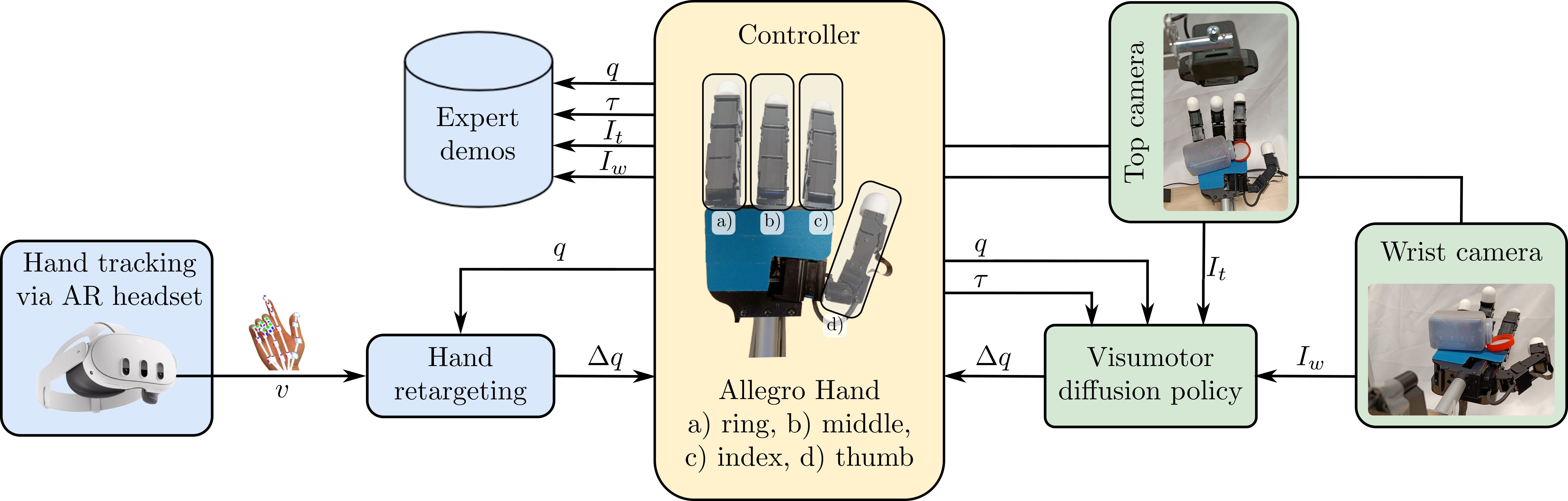}
    \caption{Overview of our system: For teleoperation (blue boxes), we obtain the operator's hand position via the Meta Quest 3 hand tracking and send the vertex positions via a Unity-ROS TCP connection. A hand retargeting node then performs inverse kinematics and motion retargeting to obtain the relative target joint positions of the Allegro Hand $\Delta q$. We save the joint position $q$, the joint effort $\tau$, and the top and wrist camera images $I_t, I_w$. 
    During autonomous operation, the trained visuomotor diffusion policy takes the Allegro Hand’s current joint position $q$, effort $\tau$, and camera images $I_t, I_w$ as input and outputs the next joint position change $\Delta q$ to execute the manipulation task.
    }
    \label{fig:sys_overview}
\end{figure*}

\section{Allegro Hand Teleoperation}

We show an overview of the system in Fig.~\ref{fig:sys_overview}. The teleoperation components are indicated in blue. In short, the operator wears an AR headset—Meta Quest 3 in our case—which provides hand tracking. This tracking data is processed by a hand retargeting node that resolves the kinematic differences between the human hand and the Allegro Hand. The final control is performed by delta joint positions. For dataset collection, we store the joint positions $q$, the joint efforts $\tau$, the top camera image $I_t$, and the wrist camera image $I_w$.

\noindent
\textbf{Hand Tracking via Meta Quest 3:} 
We deploy a custom Unity application that accesses the Meta Quest hand tracking data via the XR Hands package\footnote{\url{https://docs.unity3d.com/Packages/com.unity.xr.hands@1.1/manual/index.html}}. The vertex poses $v$ of each tracked point are transmitted through Unity’s ROS-TCP connector\footnote{\url{https://github.com/Unity-Technologies/ROS-TCP-Connector}} as a $26$-dimensional array.
For more details on Unity app development and ROS integration, see our previous works~\cite{welle2024quest2ros,lippi2024low,van2024puppeteer}.

\begin{figure}[h]
    \includegraphics[width=0.98\linewidth]{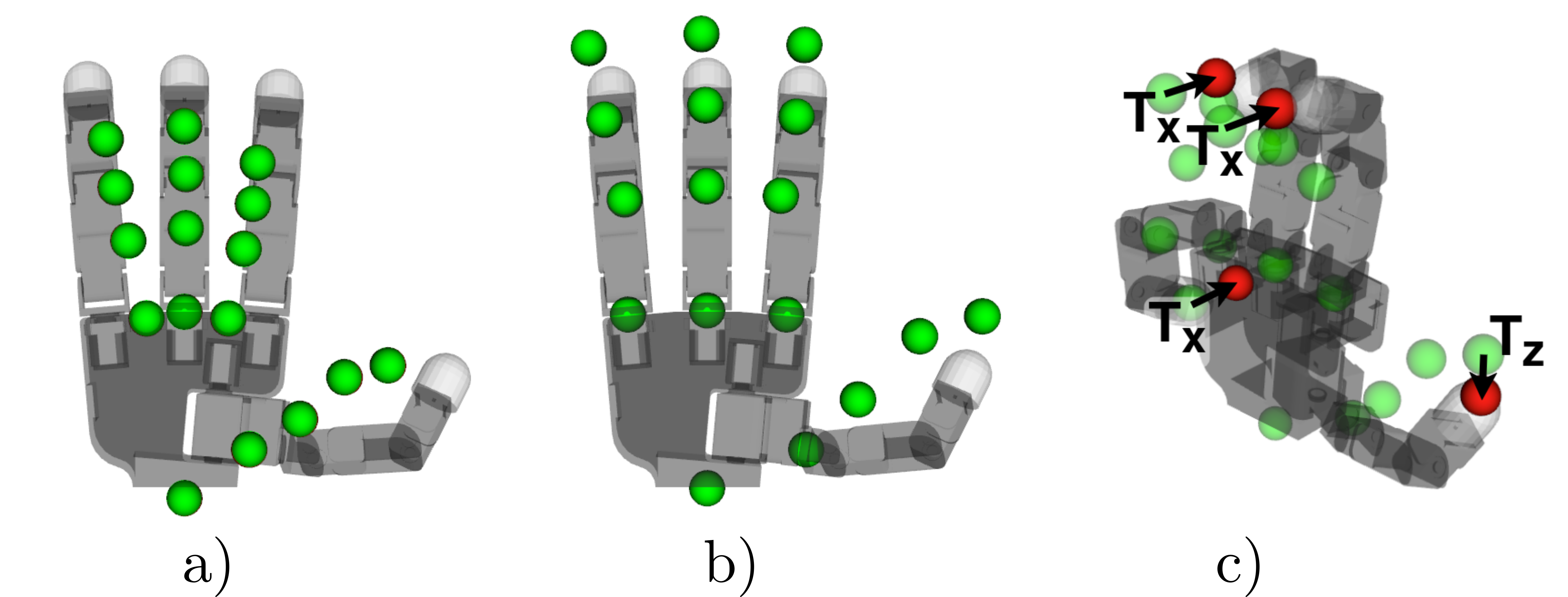} 
    \caption{Retargeting steps: (a) Initial alignment of human hand vertices (green spheres) to the Allegro Hand. (b) Scaling of finger joint lengths. (c) Final IK targets (red spheres) with additional adjustments to enhance control.
    }
    \label{fig:retargeting}
\end{figure}

\noindent
\textbf{Hand Retargeting:} 
As the kinematics of the human and Allegro hands differ substantially, we implement a series of retargeting steps to enable intuitive and precise teleoperation. These steps are illustrated in Fig.~\ref{fig:retargeting}.

\begin{enumerate}
    \renewcommand{\labelenumi}{\alph{enumi})}
    \item We remove the pinky finger and define two planes on the Allegro and human hands using the index and ring finger knuckles, along with the wrist points, to align their orientations. Additionally, we align the middle finger roots for initial translation correction (Fig.~\ref{fig:retargeting}a).
    \item We map the root positions of each finger onto their corresponding joints on the Allegro Hand, then scale the finger joint lengths using a scaling factor computed as:
    \begin{equation}
    k = \frac{ \sum_{i=1}^{n}l_i^{H} }{\sum_{i=1}^{n}l_i^{R}}
    \end{equation}
    where $n$ is the number of joints in a finger, $l_i^{H}$ is the human finger’s joint lengths obtained from the AR headset, and $l_i^{R}$ is the corresponding robot finger joint lengths. The result is shown in Fig.~\ref{fig:retargeting}b).
    \item Finally, we introduce two specific retargeting adjustments: \emph{i)} We shift the thumb fingertip by $2.3$ cm towards the wrist to maximize the Allegro Hand’s range, compensating for the human thumb’s limited reach. \emph{ii)} We shift the fingertips of the index, middle, and ring fingers $3.4$ cm towards the hand plane to facilitate full finger closure without excessive flex in operator's fingers. The final retargeted mapping is shown in Fig.~\ref{fig:retargeting}c).
\end{enumerate}

\noindent
\textbf{Individual Finger Control:}  
To maintain responsiveness, we treat each finger independently. To prevent collisions among the index, middle, and ring fingers, we fix their root joints at $0^{\deg}$ and compute inverse kinematics (IK) to reach the final retargeted fingertip positions (red points in Fig.~\ref{fig:retargeting}c)).  
The Denavit-Hartenberg (DH) parameters for these fingers are:

\begin{equation}
\text{$DH_{f}$} =
\begin{array}{cccc}
\text{Trans } X & \text{Trans } Z & \text{Rot } X & \text{Rot } Z \\
\hline
0.0 & 0.0166 & -\frac{\pi}{2} & 0 \\
0.054 & 0.0 & 0.0 & \theta_2 - \frac{\pi}{2} \\
0.0384 & 0.0 & 0.0 & \theta_3 \\
0.0437 & 0.0 & 0.0 & \theta_4
\end{array}
\end{equation}    

For the thumb:
\begin{equation}
\text{$DH_{t}$} =
\begin{array}{cccc}
\text{Trans } X & \text{Trans } Z & \text{Rot } X & \text{Rot } Z \\
\hline
0.0 & 0.0 & \frac{\pi}{2} & \theta_1 \\
0.0 & 0.0554 & -\frac{\pi}{2} & \theta_2 - \frac{\pi}{2} \\
0.0514 & 0.0 & 0.0 & \theta_3 - \frac{\pi}{2} \\
0.0593 & 0.0 & 0.0 & \theta_4
\end{array}
\end{equation}

\subsection{Expert Demonstration Collection}
A single expert operator (one of the authors) collected $300$ demonstrations of unscrewing a bottle when placed in different positions within the Allegro Hand’s palm. Demonstration recording was initiated and stopped using a fist gesture with the left hand.

\begin{figure}[h]
    \includegraphics[width=0.98\linewidth]{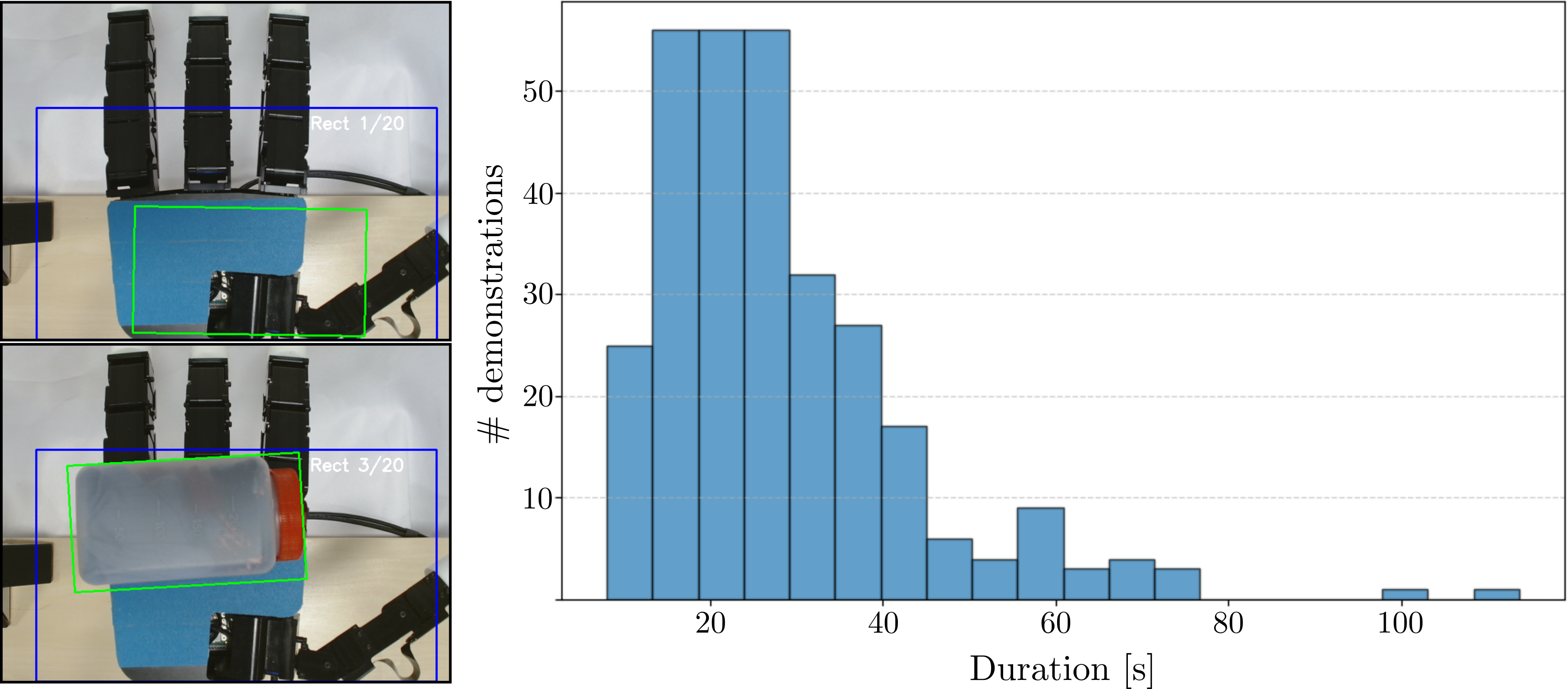} 
    \caption{Left: Examples of randomized placement prompts to ensure positional diversity. Right: Histogram of demonstration durations.
    }
    \label{fig:demos_info}
\end{figure}

To ensure diverse positional coverage, we generated random target positions within the workspace for $100$ of the demonstrations. An example of these placements and the histogram of demonstration durations is shown in Fig.~\ref{fig:demos_info}. The average demonstration length is $47.5$ seconds, resulting in a total dataset duration of $2.37$ hours, which corresponds to approximately $5$ hours of real-time execution. The operator's viewpoint while performing the demonstrations is available on the project website.

\section{Visuomotor Diffusion Policies for Complex In-Hand Manipulation Tasks}
\label{sec:visuomotor_diffusion}

In this section, we describe how we employ visuomotor diffusion policies for complex in-hand manipulation tasks. We first recap the visuomotor diffusion policy method employed~\cite{chi2023diffusionpolicy} and then describe our outlier removal paradigm, which ensures high-quality demonstration data.

\begin{figure}[t]
    \centering
    \includegraphics[width=0.98\linewidth]{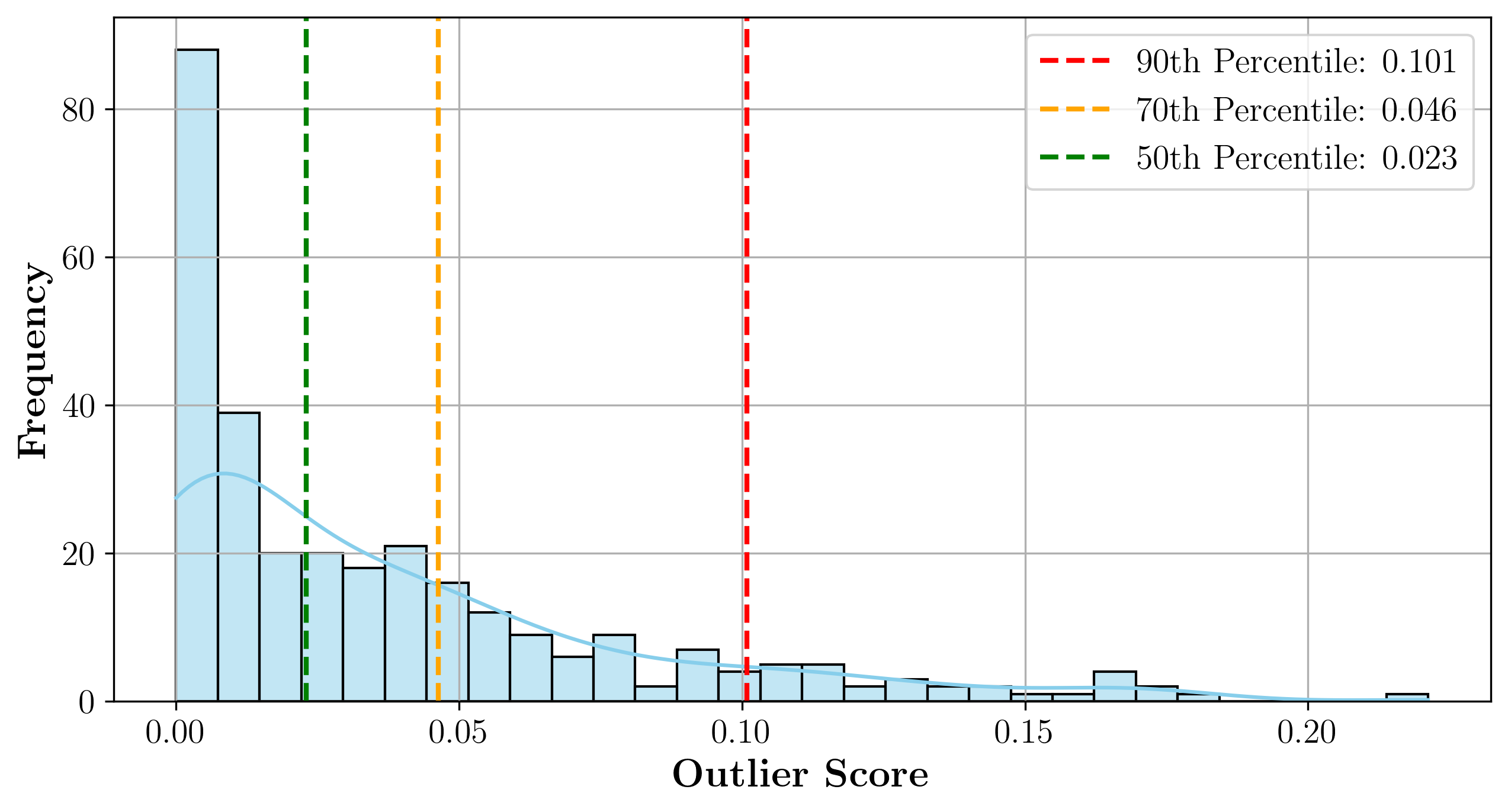}
    \caption{Distribution of outlier scores, with vertical lines marking the $90$th, $70$th, and $50$th percentiles.}
    \label{fig:out_histogram}
\end{figure}

\subsection{Visuomotor Diffusion Policy}
The Visuomotor Diffusion Policy~\cite{chi2023diffusionpolicy} formulates robot control as a conditional denoising diffusion process, iteratively refining actions instead of predicting them directly. The CNN-based version (used in this work) utilizes a 1D temporal convolutional network (CNN) to model action sequences, starting from Gaussian noise and progressively denoising through a learned noise prediction network. Feature-wise Linear Modulation (FiLM) conditions the CNN on visual inputs, ensuring responsive and temporally consistent actions. This approach efficiently captures low-frequency action patterns. 
By leveraging closed-loop action prediction, the policy continuously updates action sequences based on new observations, improving smoothness and long-horizon planning. This process ensures that the policy gradually refines actions while leveraging visual context, effectively denoising the initial action sequence to produce a feasible and coherent motion plan.

\subsection{Outlier Removal}
When collecting a large number of demonstrations, even experienced operators may produce inconsistent or suboptimal data. Furthermore, having a method to remove low-quality demonstrations as outliers enables less proficient operators to contribute to the dataset without degrading policy performance.
Our outlier removal paradigm is based on the visual observations that the policy receives. As a first step, we encode the raw images using a pre-trained ConvNeXt-Tiny model to extract meaningful feature representations. These embeddings are then clustered using HDBSCAN, which assigns outlier scores using the Global-Local Outlier Score from Hierarchies (GLOSH). This score quantifies how well a data point fits into its local density structure, helping us identify anomalous demonstrations.
Algorithm~\ref{alg:outlier_detection} outlines our outlier detection approach. We first extract features from images recorded by the top and wrist cameras. These features are then clustered using HDBSCAN, and the GLOSH outlier scores are computed for each demonstration. The final outlier ranking is obtained by averaging the scores from both cameras. For HDBSCAN, we set the $min\_cluster\_size=2$.

\begin{algorithm}
\caption{Outlier Detection using ConvNeXt-Tiny and HDBSCAN with GLOSH}
\label{alg:outlier_detection}
\begin{algorithmic}[1]
\Require $D$: Set of demonstration files, $M$: Pre-trained ConvNeXt-Tiny model
\Ensure Outlier scores for each demonstration\\
\Comment{Extract features from each demonstration}   
\For{\textbf{each} $I_t, I_w \in D$}  
    \State $top\_features \gets \text{extract\_convnext\_features}(I_t)$
    \State $wrist\_features \gets \text{extract\_convnext\_features}(I_w)$
\EndFor

\State $T \gets$ Stack all $top\_features$
\State $E \gets$ Stack all $wrist\_features$

\State \Comment{Perform clustering on feature embeddings}
\State $clustering_T \gets \text{HDBSCAN}(T)$
\State $clustering_E \gets \text{HDBSCAN}(E)$

\State \Comment{Compute GLOSH-based outlier scores}
\State $scores_T \gets$ GLOSH scores from $clustering_T$
\State $scores_E \gets$ GLOSH scores from $clustering_E$

\For{\textbf{each} $i \in \{1, ..., |D|\}$}
    \State $outlier\_score[i] \gets (scores_T[i] + scores_E[i]) / 2$
\EndFor

\State Save sorted outlier scores to file

\State \Return Outlier scores
\end{algorithmic}
\end{algorithm}

The resulting outlier histogram, with indications of the $90$th, $70$th, and $50$th percentile thresholds, is shown in Fig.~\ref{fig:out_histogram}.

From the $300$ demonstrations, we observe that a subset receives relatively high outlier scores. The advantage of our clustering-based outlier removal approach is that it captures multiple modes of the data while still effectively removing low-quality or anomalous demonstrations that could negatively impact policy performance.

\section{Experimental evaluation}
\label{sec:experiments}

We conduct real-world lid unscrewing experiments to answer the following key questions:
\begin{enumerate}
    \item How do different input modalities (cameras, joint effort) affect task performance?
    \item Does removing outliers impact task performance?
    \item What is the overall success rate of visuomotor diffusion policies for complex in-hand manipulation tasks such as unscrewing the lid of a bottle in hand?
\end{enumerate}

\subsection{Model Training and Experimental Setup}
To evaluate question \emph{1)}, we train four different policies using all $300$ demonstrations. Specifically, we train:
\begin{itemize}
    \item $\pi_{all}$, which includes all available inputs (top camera $I_t$, wrist camera $I_w$, joint effort $\tau$, and joint positions $q$).
    \item $\pi_{nt}$, which excludes the top camera.
    \item $\pi_{nw}$, which excludes the wrist camera.
    \item $\pi_{ne}$, which excludes joint effort.
\end{itemize}

Furthermore, to assess the impact of outlier removal on performance (question $2$), we train three additional policies where demonstrations above the $90$th, $70$th, and $50$th percentile of outlier scores are removed. We denote these policies as $\pi_{all}^{90}$ ,  $\pi_{all}^{70}$, and  $\pi_{all}^{50}$ respectively.
\newline

For all policies, we set:
\begin{itemize}
    \item Prediction horizon: $16$ steps
    \item Action horizon: $8$ steps
    \item Past observations: $3$ steps
    \item Training: $600$ epochs (early stopping with $10\%$ validation split and patience of $25$ epochs)
\end{itemize}

\noindent
\textbf{Experimental Setup:}  
Each policy is evaluated in $20$ real-world trials. A trial is considered successful if the policy can unscrew the lid within $2$ minutes. To ensure a fair comparison between policies, we generate $20$ novel randomized rectangular placements as positioning guides for the operator. This ensures comparable starting positions across evaluations.  

Examples of the evaluation placements are shown in Fig.~\ref{fig:demos_info}. All experimental videos ($140$) are available on the project's website.

\subsection{Experimental Results}

\begin{figure}
    \centering
    \includegraphics[width=0.98\linewidth]{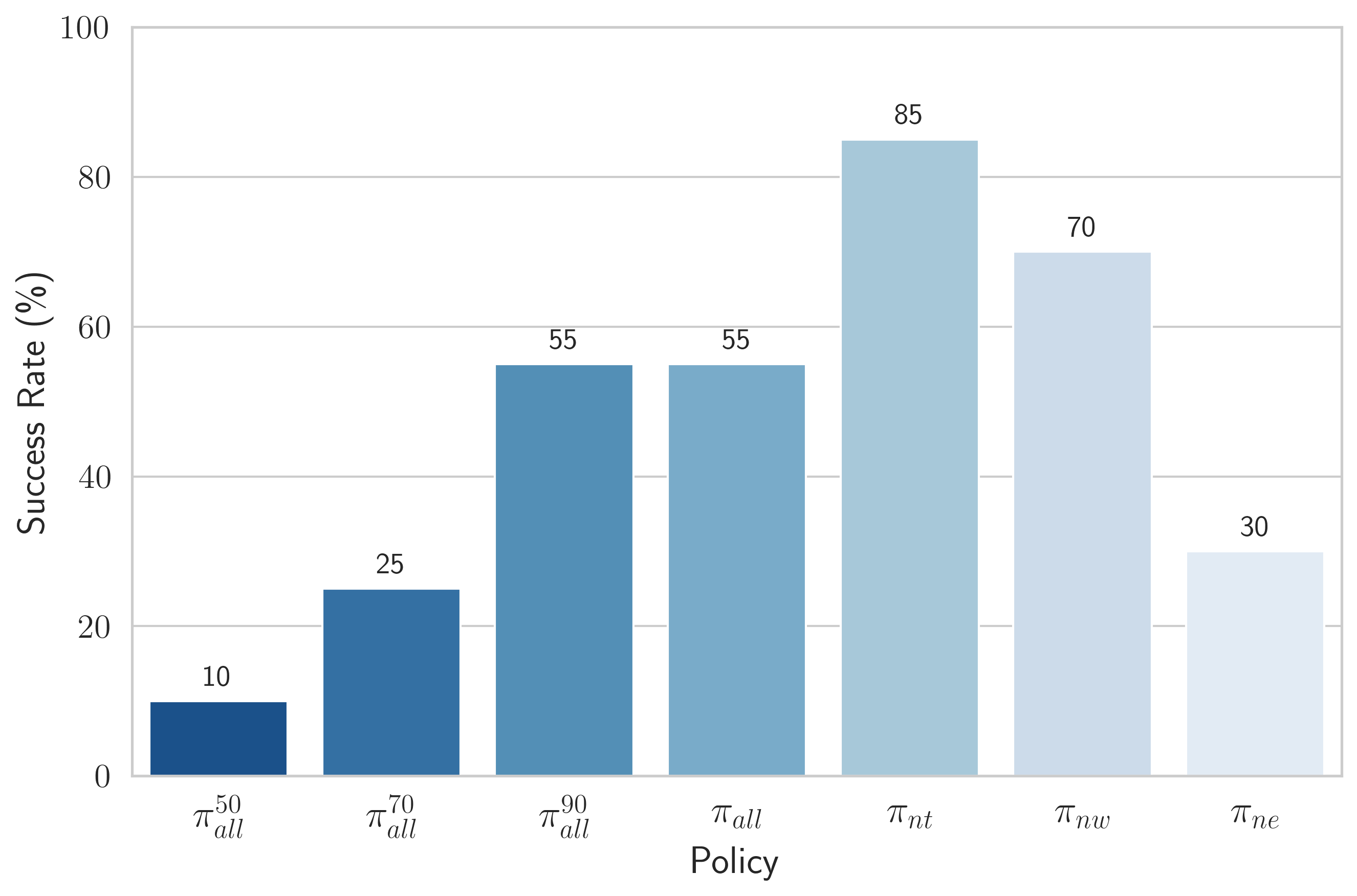}
    \caption{Success rate of the $7$ models evaluated. The best model is $\pi_{nt}$ which has the wrist camera as well as the joint position and effort as observations.}
    \label{fig:results}
\end{figure}

The results are shown in Fig.~\ref{fig:results}. The baseline policy, $\pi_{all}$, which is trained with all $300$ demonstrations and receives all available inputs (top and wrist camera $I_t, I_w$, joint position $q$, and joint effort $\tau$), achieves a $55\%$ success rate. To answer the first question regarding the effect of different input modalities on performance, we compare this result with the ablation policies $\pi_{nt}$, $\pi_{nw}$, and $\pi_{ne}$. 
We observe that removing effort information ($\pi_{ne}$) is detrimental to policy performance, yielding only a $30\%$ success rate. In particular, we note an increased number of failure cases due to the absence of feedback regarding whether the bottle is successfully grasped. This often leads to the bottle slipping out of the grasp or being mispositioned in numerous trials.
Interestingly, removing either camera input improves the policy’s performance compared to using both camera views simultaneously. This finding highlights that more camera viewpoints do not necessarily result in better performance if the information contained in a single observation is sufficient to solve the task. This observation is consistent with our previous work~\cite{ingelhag2024robotic}. 
Using only the top camera ($\pi_{nw}$) results in a $70\%$ success rate; however, relying on a top-down view
for dexterous manipulation is impractical, especially when considering deployment on mobile manipulation platforms. In contrast, the wrist-camera-only policy ($\pi_{nt}$) is much more feasible, as mobile manipulation platforms, such as humanoid robots, often already have or can be easily equipped with wrist cameras. The final performance of $85\%$ also demonstrates that the task is solved more reliably with this configuration.
When evaluating the impact of outlier removal with $\pi_{all}^{90}$, $\pi_{all}^{70}$, and $\pi_{all}^{50}$, we see that removing the top 10\% most outlier demonstrations ($\pi_{all}^{90}$) maintains policy performance despite being trained with only $270$ demonstrations instead of $300$. However, removing many more demonstrations significantly degrades performance, with success rates dropping to $25\%$ for $\pi_{all}^{70}$ and $10\%$ for $\pi_{all}^{50}$. 
This indicates that filtering out the most significant outliers does not hinder task performance; however, removing too many demonstrations leads to a substantial decline in success rate.

In summary, we are able to show that when using the wrist camera $I_w$, the joint position $q$, and the joint effort $\tau$ observation we are able to reach an overall success rate of $85\%$ for the dexterous in-hand manipulation task of unscrewing a lid from a bottle.

\section{Success and Failure Cases}

\begin{figure*}
    \centering
    \includegraphics[width=0.99\linewidth]{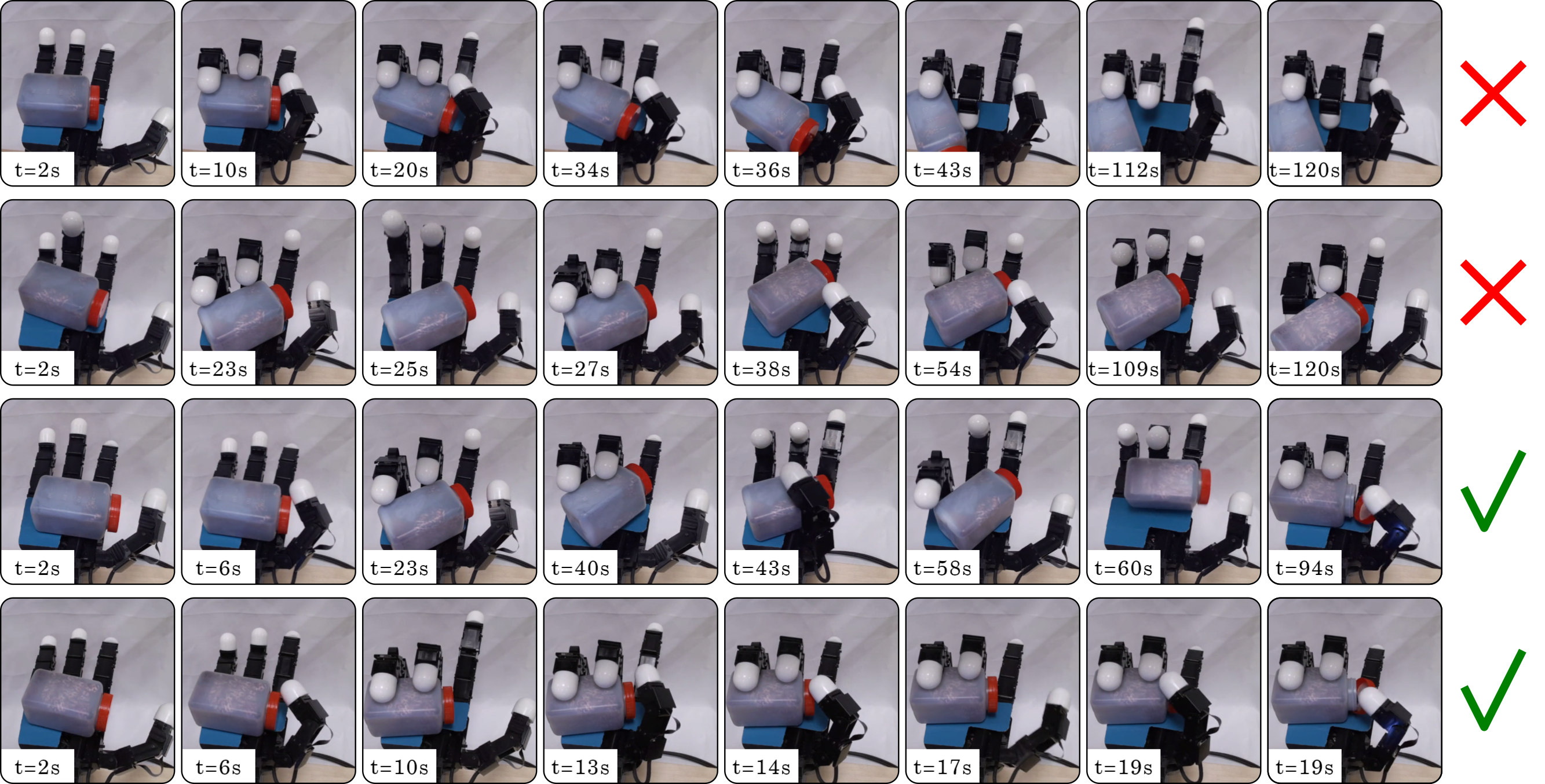}
    \caption{
    Selected examples of failure and success cases from the best-performing model, $\pi_{nt}$. The top row illustrates a failure case where the bottle was pushed out of the grasp, resulting in an unrecoverable position. The second row presents another failure case in which the policy failed to reorient the bottle correctly, preventing it from finding a valid approach angle for the thumb. The third row shows a successful case where the policy was able to recover from initial mistakes. The last row depicts the best case, where after pushing the bottle to the center, the policy quickly and efficiently completed the unscrewing task.
    }
    \label{fig:failsuc}
\end{figure*}

In this section, we analyze selected success and failure cases to better understand the limitations and potential of the current framework.
Fig.~\ref{fig:failsuc} presents two successful and two failure cases of the unscrewing task, with timestamps indicating key moments in each trial. 
In the first row, the policy failed to unscrew the bottle as it lost control of it, pushing it out of a stable grasp and resulting in an unrecoverable position. 
The second row illustrates another failure case, where the policy misaligned the bottle early in the task, preventing a stable grip from being established. This led to repeated failed attempts to reposition the object, ultimately causing the trial to terminate unsuccessfully.

The third row shows a successful trial in which the policy initially struggled to apply the correct unscrewing motion but was able to recover. After multiple adjustment attempts, the policy successfully reoriented the bottle and completed the task, albeit with a longer execution time.
Finally, the last row demonstrates the fastest successful trial recorded among the $20$ evaluations, completing the task in $19$ seconds. In this case, the policy efficiently positioned the bottle in the hand and executed a rapid and stable unscrewing motion with minimal adjustments.
Videos of all experimental results are available on the project’s website.

\section{Conclusion}

In this work, we have presented a framework for learning dexterous in-hand manipulation with multifingered hands using visuomotor diffusion policies. We introduced an intuitive and responsive teleoperation system that enables the collection of high-quality expert demonstrations via an augmented reality (AR) interface. By leveraging inverse kinematics and motion retargeting, we facilitated precise control of the Allegro Hand, allowing the collection of $300$ demonstrations for unscrewing a bottle lid.

To enhance policy training, we implemented an outlier removal approach based on HDBSCAN clustering and the Global-Local Outlier Score from Hierarchies (GLOSH), effectively filtering out low-quality demonstrations. Our experimental results demonstrate that filtering out the most significant outliers does not degrade task performance, while excessive filtering negatively impacts success rates. Furthermore, our ablation studies revealed that a combination of wrist-mounted camera observations, joint positions, and effort readings yields the best policy performance, highlighting the potential of deploying such visuomotor diffusion-based systems on mobile manipulation platforms, including humanoid robots.
Through real-world evaluations, we achieved an $85\%$ success rate using the wrist camera, joint positions, and joint effort as input modalities, demonstrating the feasibility of our approach. Our findings suggest that a wrist-mounted camera provides robust sensory feedback for dexterous tasks.
This opens up the exciting opportunity to integrate such a framework into a mobile manipulation setting.

\balance
\bibliographystyle{template/IEEEtran}
\bibliography{template/IEEEabrv,references}

\end{document}